\pdfoutput=1

\documentclass[11pt]{article}

\usepackage[]{acl}

\usepackage{times}
\usepackage{latexsym}

\usepackage[T1]{fontenc}

\usepackage[utf8]{inputenc}

\usepackage{microtype}

\usepackage{bm}
\usepackage{times}
\usepackage{latexsym}
\usepackage{amsmath,amsfonts,amssymb,bbm,epsfig,bm}
\usepackage{booktabs}
\usepackage{multirow}
\usepackage{array}
\usepackage{url}
\usepackage{tikz}
\usepackage{makecell}
\usepackage{framed}
\usepackage{pstricks}
\usepackage{xcolor}
\usepackage{enumerate}
\usepackage{arydshln}
\usepackage{algorithm}
\usepackage{subfig}
\usepackage{pifont}
\usepackage{xspace}

\title{A Two-Stream AMR-enhanced Model for Document-level \\ Event Argument Extraction}

\author{
Runxin Xu$^{1*}$, 
Peiyi Wang$^1$\thanks{\, Equal contribution.},
Tianyu Liu$^3$, Shuang Zeng$^{1,2}$,
Baobao Chang$^{1}$\thanks{\, Corresponding authors.},
Zhifang Sui$^{1}$ \\
$^1$Key Laboratory of Computational Linguistics, Peking University, MOE, China \\
$^2$School of Software and Microelectronics, Peking University, China \\
$^3$Tencent Cloud Xiaowei \\
\texttt{
 \{runxinxu,wangpeiyi9979\}@gmail.com, rogertyliu@tecent.com} \\
\texttt{
\{zengs,chbb,szf\}@pku.edu.cn
}
}

\begin{document}
\maketitle

\newcommand{\modelname}{\textsc{Tsar}\xspace}
\newcommand{\modelnamebase}{\textsc{Tsar}$_{\mathrm{base}}$\xspace}
\newcommand{\modelnamelarge}{\textsc{Tsar}$_{\mathrm{large}}$\xspace}
\newcommand{\vect}[1]{\mathbf{#1}}

\begin{abstract}
Most previous studies aim at extracting events from a single sentence, while document-level event extraction still remains under-explored.
In this paper, we focus on extracting event arguments from an entire document, which mainly faces two critical problems:
a) the long-distance dependency between trigger and arguments over sentences;
b) the distracting context towards an event in the document.
To address these issues, we propose a \textbf{T}wo-\textbf{S}tream \textbf{A}bstract meaning \textbf{R}epresentation enhanced extraction model (\textbf{\modelname}).
\modelname encodes the document from different perspectives by a two-stream encoding module, to utilize local and global information and lower the impact of distracting context.
Besides, \modelname introduces an AMR-guided interaction module to capture both intra-sentential and inter-sentential features, based on the locally and globally constructed AMR semantic graphs.
An auxiliary boundary loss is introduced to enhance the boundary information for text spans explicitly.
Extensive experiments illustrate that \modelname outperforms previous state-of-the-art by a large margin, with $2.54$ F1 and $5.13$ F1 performance gain on the public RAMS and WikiEvents datasets respectively, showing the superiority in the cross-sentence arguments extraction.
We release our code in \url{https://github.com/PKUnlp-icler/TSAR}.
\end{abstract}

\section{Introduction}
\begin{figure}
    \centering
    \includegraphics[width=\linewidth]{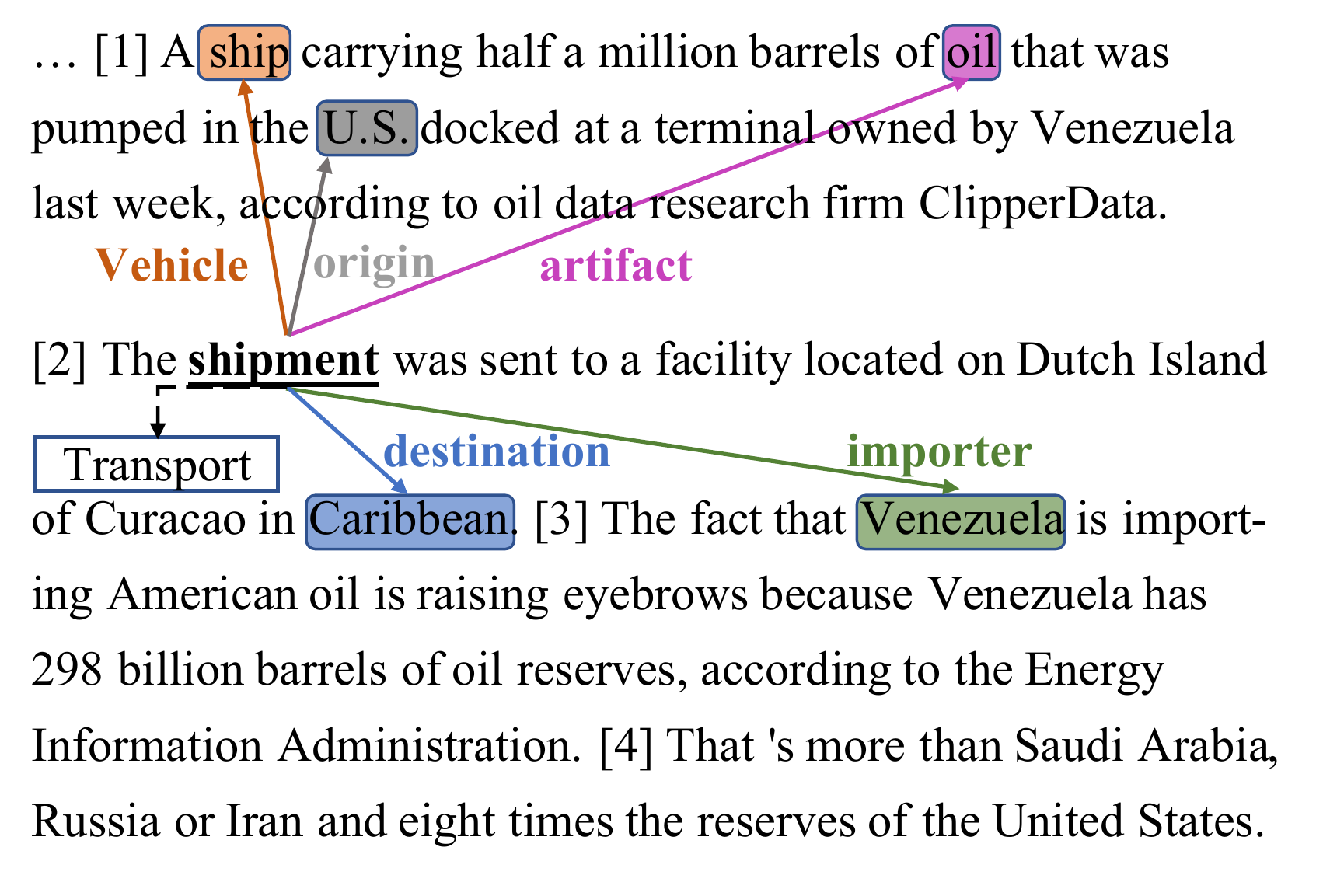}
    \caption{
    A document from RAMS dataset~\cite{rams}.
    A \emph{transport} event is triggered by \textbf{\underline{shipment}}, with five event arguments scattering across the document.
    }
\label{fig:running-example}
\end{figure}

Event Argument Extraction (EAE) aims at identifying the entities that serve as event arguments, and predicting the roles they play in the event, which is a key step for Event Extraction (EE).
It helps to transform the unstructured text into structured event knowledge that can be further utilized in recommendation systems~\cite{li-etal-2020-gaia}, dialogue systems~\cite{DBLP:conf/aime/ZhangCB20}, and so on.
Most previous studies assume that the events are only expressed by a single sentence and hence focus on sentence-level extraction~\cite{chen-etal-2015-event, liu-etal-2018-jointly, DBLP:conf/aaai/Zhou0ZWXL21}.
However, in real-life scenarios, the events are often described through a whole document consisting of multiple sentences (e.g., a news article or a financial report), which still remains under-explored.

Figure~\ref{fig:running-example} illustrates an example of document-level EAE, in which a \emph{Transport} event is triggered by \emph{shipment}.
Different from sentence-level EAE, extracting arguments out of the entire document faces two critical challenges.
(1) \textbf{Long-distance dependency} among trigger and arguments.
The arguments are usually located in different sentences from the trigger word and their distance can be quite far away. 
In Figure~\ref{fig:running-example}, while the trigger \emph{shipment} is in Sentence $2$, the \emph{vehicle}, \emph{origin}, \emph{artifact}, and \emph{importer} arguments are located in Sentence $1$ or $3$, which highly increases the extraction difficulty.
To accommodate the long-range extraction, not only intra-sentential but also inter-sentential semantics should be well modeled.
(2) \textbf{Distracting context}.
While a document naturally encompasses more context than a single sentence, some distracting context can mislead the argument extraction.
As shown in Figure~\ref{fig:running-example}, the \emph{origin} argument \emph{U.S.} can be identified more easily without Sentence $4$, which does not offer useful information for the event, but contains many place entities that can be distracting, like \emph{Saudi Arabia} and \emph{Russia or Iran}.
It remains challenging to pinpoint the useful context while discarding the distracting information.

Recently, \citet{du-cardie-2020-document} use a tagging-based method, which fails to handle nested arguments.
Instead, span-based methods predict argument roles for candidate spans~\citep{rams, two-step}. 
Some studies directly generate arguments based on sequence-to-sequence model~\citep{wikievent}.
However, how to model long-distance dependency among trigger and arguments, and how to handle distracting context explicitly, remain largely under-explored.

In this paper, to tackle the aforementioned two problems, we propose a \textbf{T}wo-\textbf{S}tream \textbf{A}M\textbf{R}-enhanced extraction model (\textbf{\modelname}).
In order to take advantage of the essential context in the document, and avoid being misled by distractions, we introduce a two-stream encoding module.
It consists of a global encoder that encodes global semantics with as much context as possible to gather adequate context information, and a local encoder that focuses on the most essential information and prudently takes in extra context.
In this way, \modelname can leverage complementary advantages of different encoding perspectives, and therefore make better use of the feasible context to benefit the extraction.
Besides, to model the long-distance dependency, we introduce an AMR-guided interaction module.
Abstract Meaning Representation ~\cite[AMR,][]{banarescu-etal-2013-abstract} graph contains rich hierarchical semantic relations among different concepts, which makes it favorable for complex event extraction.
From such a linguistic-driven angle, we turn the linear structure of the document into both global and local graph structures, followed by a graph neural network to enhance the interactions, especially for those non-local elements.
Finally, as \modelname extracts arguments in span level, where the span boundary may be ambiguous, we introduce an auxiliary boundary loss to enhance span representation with calibrated boundary.

To summarize, our contributions are three-fold.
1) We propose a two-stream encoding module for document-level EAE, which encodes the document through two different perspectives to better utilize the context.
2) We introduce an AMR-guided interaction module to facilitate the semantic interactions within the document, so that long-distance dependency can be better captured.
3) Our experiments show that \modelname outperforms the previous start-of-the-art model by large margins, with $2.54$ F1 and $5.13$ F1 improvements on public RAMS and WikiEvents datasets respectively, especially on cross-sentence event arguments extraction.

\section{Related Work}
\begin{figure*}
    \centering
    \includegraphics[width=\linewidth]{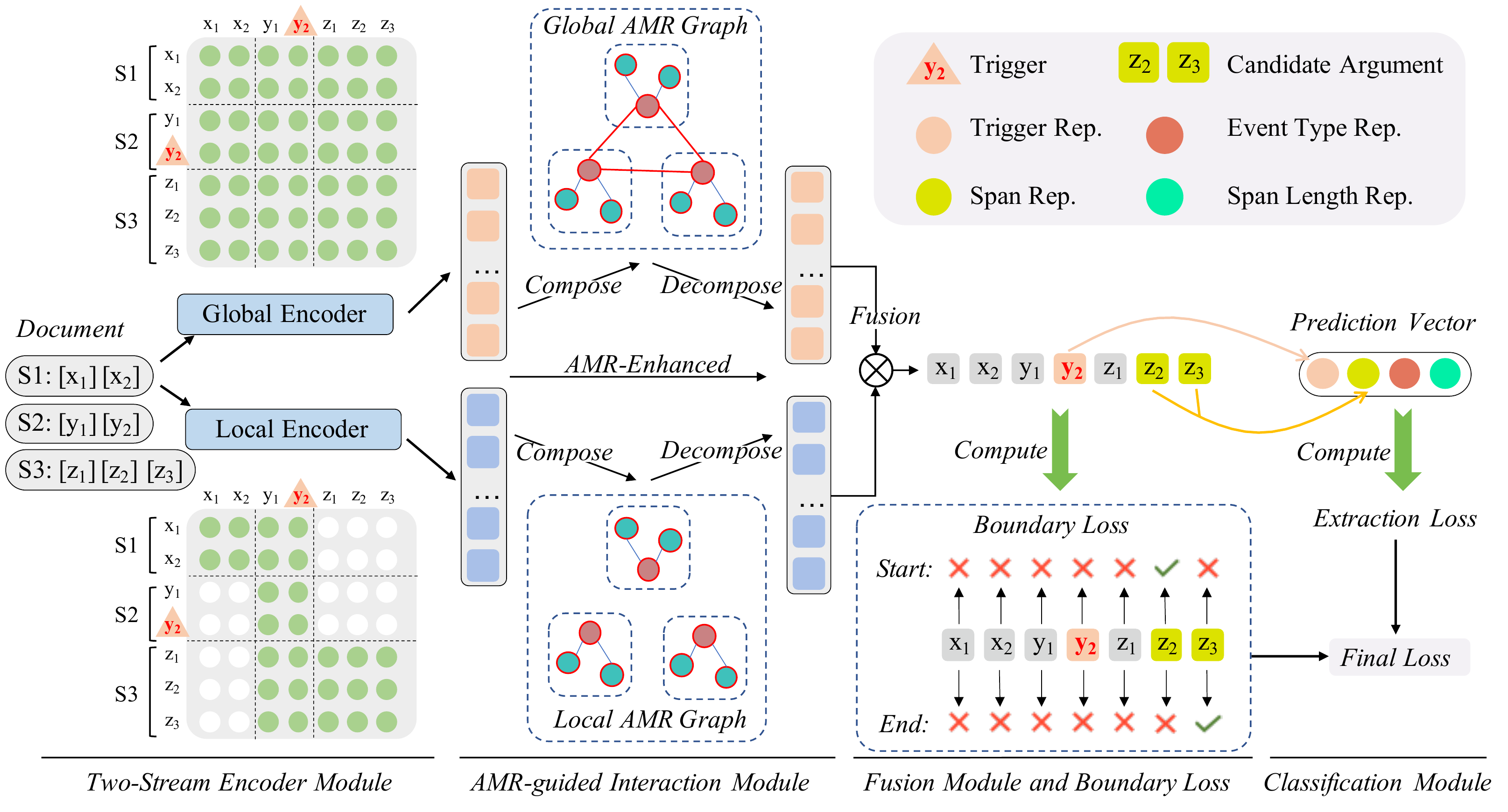}
    \caption{
    \textbf{Overview of our \modelname}.
    Firstly, taking an entire document as input, \modelname first encodes the document by the two-stream encoding module, where the global and local encoders with different attention reception fields are used to capture the context in different scopes.
    Then the AMR-guided interaction module constructs global AMR graphs and local ones to stimulate the interactions among concepts in the document, especially those far away from each other, based on graph neural network.
    Next, the information fusion module fuses the two-stream representations, and also strengthens the boundary information through a boundary loss.
    Finally, the classification module makes predictions for candidate spans.
    For conciseness, we assume the document has three sentences, $S_1$, $S_2$, $S_3$, and the event is triggered by $y_2$ with $[z_2, z_3]$ being a candidate argument span.
    }
\label{fig:model}
\end{figure*}

\subsection{Sentence-level Event Extraction}
Previous studies mainly focus on sentence-level event extraction.
\citet{li-etal-2014-constructing} and \citet{judea-strube-2016-incremental} use handcrafted features to extract events from the sentence. \citet{chen-etal-2015-event} firstly propose a neural pipeline model to extract events, while \citet{nguyen-etal-2016-joint} utilize a joint model to mitigate error propagation.
To better model the interactions among  words, \citet{liu-etal-2018-jointly, yan-etal-2019-event, ma-etal-2020-resource} make use of the dependency tree, and \citet{wadden-etal-2019-entity} enumerates all possible spans and propagate information in the span graph.
Data augmentation is also considered~\citep{yang-etal-2019-exploring-pre}.
Moreover, some works try to reformulate the event extraction task as other tasks.
For example, \citet{du-cardie-2020-event} and \citet{DBLP:conf/aaai/Zhou0ZWXL21} cast event extraction as question answering, and \citet{DBLP:journals/corr/abs-2107-00189} model it as a sequence-to-sequence task.
However, all of these models can only extract events from a single sentence.
Thus, they fail to handle the much more common cases, where event arguments usually spread over multiple sentences within the document.

\subsection{Document-level Event Extraction}

In order to extract events from a whole piece of article with multiple sentences, document-level event extraction has attracted more and more attention recently.
\citet{yang-mitchell-2016-joint} utilize well-defined features to extract arguments across sentences, while most recent methods are based on neural networks.
Some studies first identify entities in the document, followed by assigning these entities as specific argument roles~\citep{yang-etal-2018-dcfee, zheng-etal-2019-doc2edag, xu-etal-2021-git}.
Differently, some studies try to jointly extract entities and argument roles simultaneously, which can be further divided into tagging-based and span-based methods.
Tagging-based methods directly conduct sequence labeling for each token in the document with BIO-schema~\citep{du-cardie-2020-document, DBLP:conf/pakdd/VeysehDTMWJKCN21}, while span-based methods predict the argument role for candidate text spans which usually have a maximum length limitation~\citep{rams, two-step}.
Another line of studies reformulate the task as a sequence-to-sequence task~\citep{du-etal-2021-grit,du-etal-2021-template,wikievent}, or machine reading comprehension task~\citep{wei-etal-2021-trigger}.

As a span-based method, \modelname is different from prior methods that simply encode it as a long sentence.
Instead, \modelname introduces a two-stream encoding module and AMR-guided interactions module to model intra-sentential and inter-sentential semantics, along with an auxiliary boundary loss to enhance span boundary information.

\section{Task Formulation}
Following~\citet{rams}, we formulate doc-level event argument extraction as follows.
We define that a document $\mathcal{D}$ consists of $N$ sentences, and a sentence is comprised of a sequence of words, i.e., $\mathcal{D}=\left \{ w_1, w_2, \dots, w_{\left | \mathcal{D} \right | } \right \}$, and $\mathrm{SEN}\left ( w_i \right ) \in \left [1, N\right ]$ refers to the sentence that $w_i$ belongs to.
We also define the event types set $\mathcal{E}$ and the corresponding argument roles set $\mathcal{R}_e$ for each event type $e\in\mathcal{E}$.
Then, given a document $\mathcal{D}$ and the trigger $t\in \mathcal{D}$ triggering the event type $e\in\mathcal{E}$, the task aims to detect all $(r,s)$ pairs for the event, where $r \in \mathcal{R}_e$ is an argument role for the event type $e$, and $s \subseteq  \mathcal{D}$ is a contiguous text span in the document.

\section{Methodology}
Figure~\ref{fig:model} shows the overall architecture of our model \modelname. 
The document is fed into the two-stream encoding module, followed by the AMR-guided interaction module to derive both global and local contextualized representations.
The information fusion module fuses these two-stream representations, and the classification module finally predicts argument roles for candidate spans.

\subsection{Two-Stream Encoding Module}

Although more context is provided by the document, it also inevitably introduces irrelevant and distracting information towards the event.
These noise signals can be harmful to the argument extraction as shown in Figure~\ref{fig:running-example}.
To capture useful information and filter distracting one, we propose a two-stream encoding module, consisting of a global encoder that is aware of all context, and a local encoder that only prudently focuses on the most essential information.
Therefore, we can leverage their complementary advantages to make better use of the context information.

Specifically, the global and local encoders share the same Transformer-based pre-trained language model such as BERT.
By controlling the reception field of the words in the self-attention module, we can encode the document from different perspectives.
In the global encoder, the attention technique is the same as the traditional Transformer:
\begin{equation*}
\resizebox{\linewidth}{!}{$
\mathrm{Attention}^G \left (Q,K,V \right ) = \mathrm{softmax} \left (\frac{QK^\top}{\sqrt{d_m}} \right )V
$
}
\label{eq:global-encoder}
\end{equation*}
where $Q$, $K$, $V$ refers to query, key, and value matrix, and $d_m$ is the model dimension.
However, in the local encoder, we introduce a mask matrix $M$, such that tokens can only attend to the sentence itself and the sentence where the trigger locates, to avoid redundant distracting information:
\begin{equation*}
\resizebox{\linewidth}{!}{$
\mathrm{Attention}^L \left (Q,K,V 
\right ) = \mathrm{softmax} \left (\frac{QK^\top+M}{\sqrt{d_m}} \right )V
$
}
\label{eq:local-encoder}
\end{equation*}
\begin{equation*}
\resizebox{\linewidth}{!}{ $
M_{ij} =\left\{
\begin{array}{ll}
0, &  \mathrm{SEN}\left (w_j \right ) \in \left \{\mathrm{SEN} \left (w_i \right ), \mathrm{SEN}\left (t \right ) \right \} \\
-\infty, & Otherwise
\end{array} \right.
$
}
\end{equation*}
where $\mathrm{SEN}\left ( w_i \right )$ is the sentence that the word $w_i$ belongs to, and $t$ refers to the trigger of the event.

Hence, we encode the document with two different streams, a global encoder $\mathrm{Encoder}^G$ and a local encoder $\mathrm{Encoder}^L$, finally deriving two representations, $Z^G$ and $Z^L$:

\begin{equation*}
\resizebox{\linewidth}{!}{$
\begin{aligned} 
Z^G &= \left [z_1^G, z_2^G, \dots, z^G_{\left | \mathcal{D} \right | } \right ] = \mathrm{Encoder}^G \left (\left [w_1, w_2, \dots, w_{\left | \mathcal{D} \right | } \right ] \right ) \\
Z^L &= \left [z_1^L, z_2^L, \dots, z^L_{\left | \mathcal{D} \right | } \right ] = \mathrm{Encoder}^L \left ( \left [w_1, w_2, \dots, w_{\left | \mathcal{D} \right | } \right ] \right )
\end{aligned}
$
}
\end{equation*}

\subsection{AMR-Guided Interaction Module}
\begin{figure}[t]
    \centering
    \includegraphics[width=\linewidth]{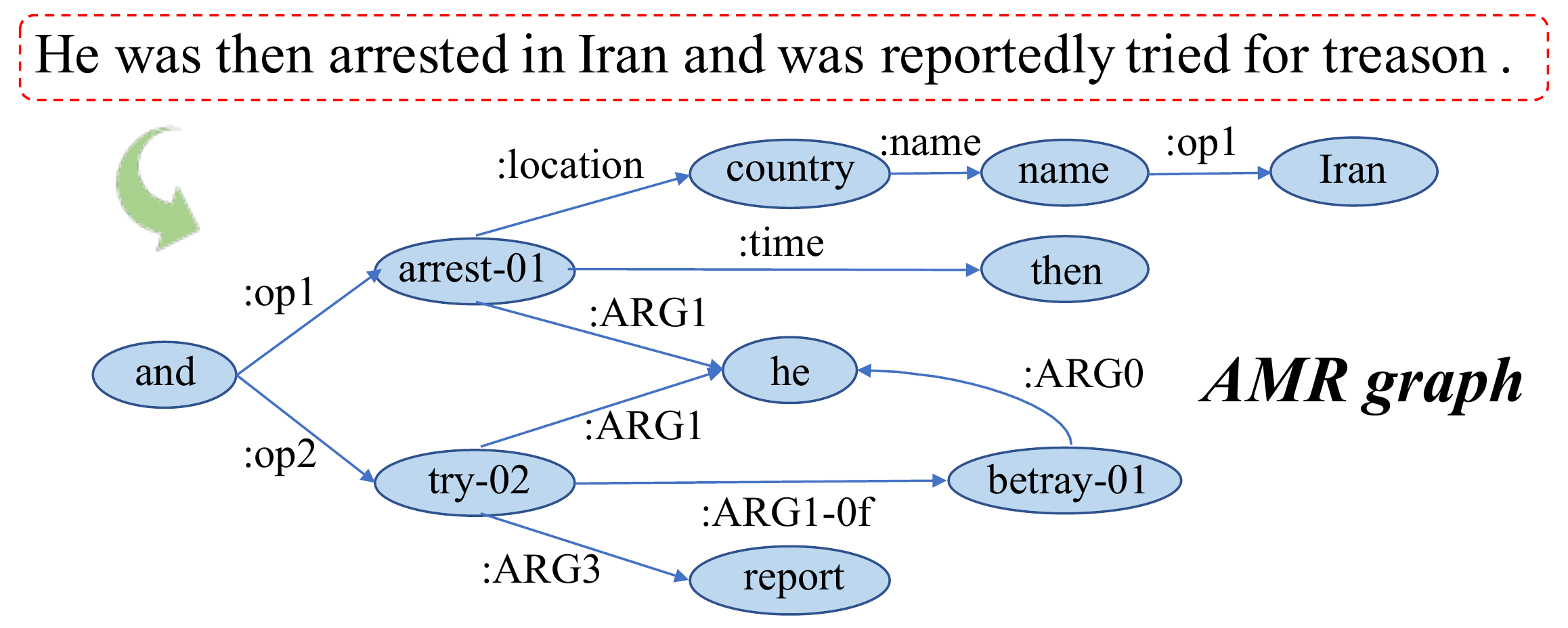}
    \caption{
    The AMR graph provides abstract and logical semantic information, where the nodes denote the concepts and the edges refer to different relation types.
    The corresponding text spans for nodes are omitted.
    }
\label{fig:amr}
\end{figure}

One key challenge to extract arguments from the document is to capture the intra-sentential and inter-sentential features.
Therefore, we propose an AMR-guided  interaction module that adopts Abstract Meaning Representation ~\cite[AMR,][]{banarescu-etal-2013-abstract} graph to provide rich semantic structure to facilitate the interactions among concepts, which also offers logical meanings of the document from a linguistic-driven perspective to benefit the language understanding.

AMR semantic graph models the meaning representations of a sentence as a rooted, directed, labeled graph.
Concretely, with an AMR parser, a natural sentence can be parsed into an AMR graph $G=(V, E)$.
The node $v=(a,b) \in V$ represents a concept that corresponds to the span ranging from $w_a$ to $w_b$ in the origin sentence, while the edge represents a specific AMR relation (detail in Appendix~\ref{sec:appendix-amr}).
Thus, AMR focuses on semantic relations rather than syntactic ones, which is more high-level and beneficial to event understanding, and the structures are more close to the event trigger-arguments  structures.
For example, Figure~\ref{fig:amr} demonstrates how a sentence is parsed into an AMR semantic graph.
As event arguments play essential roles in the text, most of them would be involved, if not all, in the AMR graphs ($90\%$ and $88\%$ arguments in RAMS and WikiEvents datasets).
We use the state-of-the-art AMR parser~\citet{fernandez-astudillo-etal-2020-transition}, which achieves satisfactory results (up to $81.3$ Smatch on AMR2.0 data) for downstream application.
As the number of AMR relation types is large, which results in too many demanded parameters, we also follow~\citet{zhangzixuan} to cluster the relation types into main categories.
More details can be found in Appendix~\ref{sec:appendix-amr}.

The AMR-guided interaction module is attached after the global and local encoders as shown in Figure~\ref{fig:model}.
We use the AMR graphs as skeletons for information interactions, under a \textit{composition, interaction, and decomposition} paradigm.

From the local perspective, we construct AMR graphs for each sentence in the document, and they are isolated from each other.
For initialization, the vector representation of node $u=(a_u, b_u)$ is \underline{\textit{\textbf{composed}}} by averaging the local representations of its corresponding text span:
\begin{equation*}
    h^0_u = \frac{1}{\left | b_u-a_u+1 \right | } \sum_{i=a_u}^{b_u} z^L_i
\end{equation*}

Similar to \citet{zeng-etal-2020-double}, we then use $L$-layer stacked Graph Convolution Network \citep{kipf2017semi} to model the \underline{\textit{\textbf{interactions}}} among different concept nodes through edges with different relation types.
Given node $u$ at the $l$-th layer, the information interaction and aggregation operation is defined as follows:
\begin{equation*}
\resizebox{\linewidth}{!}{ $
       h_{u}^{(l + 1)} = \mathrm{ReLU} \left(\sum_{k\in\mathcal{K}}\sum_{v\in\mathcal{N}_k(u) \bigcup \{u\}} \frac{1}{c_{u,k}} W^{(l)}_k h_{v}^{(l)} \right)
$
}
\end{equation*}
where $\mathcal{K}$ denotes different relation types, $\mathcal{N}_k(u)$ denotes the neighbors for $u$ connected with $k$-th relation types and $c_{u,k}$ is a normalization constant.
Besides, $W^{(l)}_k\in \mathbb{R}^{d_m \times d_m}$ is a trainable parameter.

Finally, we concatenate vectors in all layers and derive the final node representation by $h_u= W_1[h_{u}^{0}; h_{u}^{1}; \ldots; h_{u}^{L}] \in \mathbb{R}^{d_{m}}$.
Then $h_u$ is  \underline{\textit{\textbf{decomposed}}} into the local representations of corresponding words, followed by token-wise aggregation, where $\mathbb{I}(\cdot)$ refers to the indication function:
\begin{equation*}
    \widetilde{h}^L_i = z^L_i + \frac{\sum_{u} \mathbb{I}(a_u<=i \wedge b_u>=i)h_u}{\sum_{u}\mathbb{I}(a_u<=i \wedge b_u>=i)} 
\end{equation*}

From the global perspective, we first construct the global AMR graphs by fully connecting the root nodes of AMR graphs of different sentences, since the root nodes contain the core semantics according to the AMR core-semantic principle~\citep{cai-lam-2019-core}
\footnote{We find more elaborate methods yield no further improvements, so we adopt this simple connection paradigm.}.
Then similar graph-based interaction methods are used to obtain the AMR-enhanced global representations $\widetilde{h}^G_i$, but based on global AMR graphs instead.
In this way, the inter-sentential information can flow through the sentence boundaries, and therefore long-distance dependency can also be better captured.

\subsection{Information Fusion Module}

In the information fusion module, we fuse the global representations $\widetilde{H}^G = \left [\widetilde{h}_1^G, \widetilde{h}_2^G, \dots, \widetilde{h}^G_{\left | \mathcal{D} \right | } \right ]$ and local representations $\widetilde{H}^L = \left [\widetilde{h}_1^L, \widetilde{h}_2^L, \dots, \widetilde{h}^L_{\left | \mathcal{D} \right | } \right ]$, to construct the final vector representations for the candidate spans.

In detail, we use a gated fusion to control how much information is incorporated from the two-stream representations.
Given $\widetilde{h}_i^G$ and $\widetilde{h}_i^L$, we calculate the gate vector $g_i$ with trainable parameters $W_2$ and $W_3$, $g_i = \mathrm{sigmoid}(W_2\widetilde{h}_i^G+W_3\widetilde{h}_i^L+b)$.
Then we derive the fused representations $\widetilde{h}_i$:
\begin{equation*}
    \widetilde{h}_i =  g_i \odot \widetilde{h}_i^G+(1-g_i) \odot \widetilde{h}_i^L
\end{equation*}

For a candidate text span ranging from $w_i$ to $w_j$, its fused representation consists of the start representation $\widetilde{h}_{i}^{start}$, the end representation $\widetilde{h}_{j}^{end}$ and the average pooling of the hidden state of the span with $W_{span} \in \mathbb{R}^{ d_m \times (3 \times d_m)}$:
\begin{equation*}
s_{i:j} = W_{span}\left [ \widetilde{h}_i^{start};\widetilde{h}_i^{end}; \frac{1}{j-i+1} \sum_{k=i}^{j} \widetilde{h}_k \right ]
\end{equation*}
where $\widetilde{h}_{i}^{start} = W_{s}\widetilde{h}_i$ and $\widetilde{h}_{i}^{end} = W_{e}\widetilde{h}_i$.

Since we extract arguments in span level, whose boundary may be ambiguous, we introduce an auxiliary boundary loss to enhance boundary information for the $\widetilde{h}_{i}^{start}$ and $\widetilde{h}_{i}^{end}$.
In detail, we predict whether the word $w_i$ is the first or last word of a golden argument span with token-wise classifiers.
We use a linear transformation followed by a sigmoid function, to derive the probability of the word $w_i$ being the first or last word of a golden argument span, i.e., $P^s_i$ and $P^e_i$.
\begin{equation*}
\resizebox{\linewidth}{!}{ $
P^s_i = \mathrm{sigmoid}\left ( W_4 \widetilde{h}_i^{start} \right ) 
,
P^e_i = \mathrm{sigmoid}\left ( W_5 \widetilde{h}_i^{end} \right ) 
$
}
\end{equation*}
Finally, the boundary loss is defined as the following cross-entropy losses of detecting the start and end position.
\begin{equation}
\begin{aligned} 
\mathcal{L}_{b} = - \sum_{i=1}^{\left | \mathcal{D} \right | }
[  y_i^{s} \mathrm{log}P_i^s + \left (1-y_i^{s} \right )\mathrm{log} \left (1-P_i^s \right ) \\
 +y_i^{e}\mathrm{log}P_i^e + \left (1-y_i^{e} \right )\mathrm{log}\left (1-P_i^e \right ) ] 
\end{aligned} 
\end{equation}
where, $y_i^s$ and $y_i^e$ denote the golden labels.
In this way, we introduce an explicit supervision signal to inject boundary information of the start and end representation of an span, which is shown to be necessary and important to the extraction in our exploring experiments.

\subsection{Classification Module}

In the classification module, we predict what argument role the candidate span plays, or it does not belong to any specific argument roles.
Besides the span representation $s_{i:j}$, we also consider the trigger, event type, and the length of the span.
Specifically, we concatenate the following representations to obtain the final prediction vector $I_{i:j}$:
1) the trigger representation $\widetilde{h}_t$, and the span representation $s_{i:j}$, with their absolute difference $\left | \widetilde{h}_t-s_{i:j} \right |$, and element-wise multiplication, $\widetilde{h}_t \odot s_{i:j}$; 
2) the embedding of the event type $\mathrm{E}_{type}$.
3) the embedding of the span length $\mathrm{E}_{len}$;

\begin{equation*}
 I_{i:j} = \left [ \widetilde{h}_t; s_{i:j};  \left | \widetilde{h}_t-s_{i:j} \right |; \widetilde{h}_t \odot s_{i:j};
\mathrm{E}_{type}; \mathrm{E}_{len} \right ] 
\end{equation*}
We then use the cross entropy $\mathcal{L}_{c}$ as loss function:
\begin{equation}
\mathcal{L}_{c} = - \sum_{i=1}^{\left | \mathcal{D} \right | } \sum_{j=i}^{\left | \mathcal{D} \right | }
y_{i:j} \mathrm{log} P \left (r_{i:j} = y_{i:j} \right )
\end{equation}
where $y_{i:j}$ is the golden argument role, and $P\left (r_{i:j} \right)$ is derived by a feed-forward network based on $I_{i:j}$.

Finally, we train the model in an end-to-end way with the final loss function $\mathcal{L} = \mathcal{L}_{c} + \lambda\mathcal{L}_{b}$ with hyperparameter $\lambda$.

\section{Experiments}
\subsection{Datasets}
We evaluate our model on two public document-level event argument extraction datasets, RAMS v$1.0$~\cite{rams} and WikiEvents~\cite{wikievent}.
RAMS contains $9,124$ human-annotated examples, with $139$ event types and $65$ kinds of argument roles, and more than $21$k arguments.
WikiEvents is another human-annotated dataset, with $50$ event types and $59$ event argument roles, and more than $3.9$k events.
We follow the official train/dev/test split for RAMS and WikiEvents datasets, and use the evaluation script provided by ~\citet{rams} to evaluate the performance.
The detailed data statistics of RAMS and WikiEvents datasets are shown in Table~\ref{table:datasets}.

\begin{table}[t]
\centering
\scalebox{0.8}{
    \begin{tabular}{lcccc}
    \toprule
    \bf Dataset & \bf Split & \bf \# Doc. & \bf \# Event & \bf \# Argument \\
    \midrule
    \multirow{3}{*}{RAMS} & Train & 3,194 & 7,329 & 17,026 \\
    ~ & Dev & 399 & 924 & 2,188 \\
    ~ & Test & 400 & 871 & 2,023 \\
    \midrule
    \multirow{3}{*}{WikiEvents} & Train & 206 & 3,241 & 4,542 \\
    ~ & Dev & 20 & 345 & 428 \\
    ~ & Test & 20 & 365 & 566 \\
    \bottomrule
    \end{tabular}
}
\caption{
Statistics of RAMS and WikiEvents datasets.
}
\label{table:datasets}
\end{table}

\subsection{Experiment Setups and  Metrics}

In our implementation, we use BERT$_{\mathrm{base}}$~\cite{bert} and RoBERTa$_{\mathrm{large}}$~\cite{roberta} as our backbone encoder for \modelname, with global and local encoders sharing parameters.
Detailed hyperparameters are listed in Appendix~\ref{sec:appendix-hyperparameters}.

Following~\citet{two-step}, we report the Span F1 and Head F1 for RAMS dataset.
Span F1 requires the predicted argument spans to fully match the golden ones, while Head F1 relaxes the constraint and evaluates solely on the head word of the argument span.
The head word of a span is defined as the word that has the smallest arc distance to the root in the dependency tree.
In addition, following~\citet{wikievent}, we report the Head F1 and Coref F1 scores for WikiEvents dataset.
The model is given full credit in Coref F1 if the extracted argument is coreferential with the golden argument as used by~\citet{coref}.

\subsection{Main Results}
\begin{table}[t]
\centering
\scalebox{0.78}{
    \begin{tabular}{lcccc}
    \toprule
    \multirow{2}{*}{\bf Method} & \multicolumn{2}{c}{\bf Dev} & 
    \multicolumn{2}{c}{\bf Test} \\
    \cmidrule(lr){2-3} \cmidrule(lr){4-5}
    ~ & Span F1 & Head F1 & Span F1 & Head F1 \\
    \midrule
    BERT-CRF & 38.1 & 45.7 & 39.3 & 47.1 \\
    BERF-CRF$_{\mathrm{TCD}}$ & 39.2 & 46.7 & 40.5 & 48.0 \\
    Two-Step & 38.9 & 46.4 & 40.1 & 47.7 \\
    Two-Step$_{\mathrm{TCD}}$ & 40.3 & 48.0 & 41.8 & 49.7 \\
    FEAE & - & - & 47.40 & - \\
    \modelnamebase (Ours) & \bf 45.23 & \bf 51.70 & \bf 48.06 & \bf 55.04 \\
    \midrule
    \midrule
    BART-Gen & - & - & 48.64 & 57.32 \\
    \modelnamelarge (Ours) & \bf 49.23 & \bf 56.76 & \bf 51.18 & \bf 58.53 \\
    \bottomrule
    \end{tabular}
}
\caption{
\textbf{Comparison between \modelname and other methods on RAMS dataset}.
Models above the double line are based on BERT$_{\mathrm{base}}$.
\modelname consistently outperforms others on Span F1 and Head F1.
Compared with BART-Gen, \modelname improves $2.54$ Span F1 in the test set.
}
\label{table:main-rams}
\end{table}

We compare \modelname with the following baselines.
1) \textbf{BERT-CRF}~\cite{seq} is a tagging-based method, which adopts a BERT-based BIO-styled sequence labeling model.
2) \textbf{Two-Step}~\cite{two-step} is a span-based method, which first identifies the head word of possible argument span, and then extends to the full span.
\textbf{BERT-CRF$_{\mathrm{TCD}}$} and \textbf{Two-Step$_{\mathrm{TCD}}$} refers to adopting Type-Constraint Decoding mechanism as used in~\cite{rams}.
3) \textbf{FEAE}~\cite{wei-etal-2021-trigger}, Frame-aware Event Argument Extraction, is a concurrent work based on question answering.
4) \textbf{BERT-QA}~\cite{bert-qa} is also a QA-based model. BERT-QA and BERT-QA-Doc extract run on sentence-level and document-level, respectively.
5) \textbf{BART-Gen}~\cite{wikievent} formulate the task as a sequence-to-sequence task and uses BART$_{\mathrm{large}}$~\cite{bart} to generate corresponding arguments in a predefined format.

Table~\ref{table:main-rams} illustrates the results in both dev and test set on RAMS dataset.
As is shown, among models based on BERT$_{\mathrm{base}}$, \modelname outperforms other previous methods.
For example, \modelname yields an improvement of $4.93\sim7.13$ Span F1 and $3.70\sim6.00$ Head F1 compared with the previous method in the dev set, and up to $8.76$ Span F1 in the test set.
Besides, among models based on large pre-trained language models, \modelname  outperforms BART-Gen by $2.54$ Span F1 and $1.21$ Head F1\footnote{We use \modelnamelarge based on RoBERTa$_{\mathrm{large}}$ to compare with BART-Gen based on BART$_{\mathrm{large}}$, as they are pre-trained on the same corpus with the same batch size and training steps.}.
These results suggest that encoding the document in a two-stream way, and introducing AMR graphs to facilitate interactions, is beneficial to capturing intra-sentential and inter-sentential features, and thus improves the performance.

\begin{table}[t]
\centering
\scalebox{0.75}{
    \begin{tabular}{lcccc}
    \toprule
    \multirow{2}{*}{\bf Method} & \multicolumn{2}{c}{\bf Arg Identification} & 
    \multicolumn{2}{c}{\bf Arg Classification} \\
    \cmidrule(lr){2-3} \cmidrule(lr){4-5}
    ~ & Head F1 & Coref F1 & Head F1 & Coref F1 \\
    \midrule
    BERT-CRF & 69.83 & 72.24 & 54.48 & 56.72 \\
    BERT-QA & 61.05 & 64.59 & 56.16 & 59.36 \\
    BERT-QA-Doc & 39.15 & 51.25 & 34.77 & 45.96 \\
    \modelnamebase (Ours) & \bf 75.52 & \bf 73.17 & \bf 68.11 & \bf 66.31 \\
    \midrule
    \midrule
    BART-Gen & 71.75 & 72.29 & 64.57 & 65.11 \\
    \modelnamelarge (Ours) & \bf 76.62 & \bf 75.52 & \bf 69.70 & \bf 68.79 \\
    \bottomrule
    \end{tabular}
}
\caption{
\textbf{Comparison between \modelname and other methods on WikiEvents dataset}.
Models above the double line are based on BERT$_{\mathrm{base}}$.
\modelname yields evident improvements in argument identification and classification sub-tasks.
Compared with BART-Gen, \modelname improves Head F1 in argument classification by $5.13$ score.
}
\label{table:main-wikievent}
\end{table}

Moreover, we follow~\citet{wikievent} to evaluate both argument identification and argument classification, and report the Head F1 and Coref F1.
Identification requires the model to correctly detect the argument span boundary, while classification has to further correctly predict its argument role.
As illustrated in Table~\ref{table:main-wikievent}, \modelname consistently outperforms others in both tasks.
Compared with BART-Gen, \modelname improves up to $4.87$/$3.23$ Head/Coref F1 for argument identification, and $5.13$/$3.68$ Head/Coref F1 for argument classification.
Similar results also appear among models based on BERT$_{\mathrm{base}}$, with $5.69\sim36.37$ and $11.95\sim33.34$ Head F1 improvement for identification and classification.
These results show that \modelname is superior to other methods in not only detecting the boundary of argument spans, but also predicting their roles.

\section{Analysis}
\subsection{Cross-sentence Argument Extraction}
\begin{table}[t]
\centering
\scalebox{0.88}{
    \begin{tabular}{lccccc}
    \toprule
    \bf Method & d=-2 & d=-1 & d=0 & d=1 & d=2 \\
    \midrule
    BERT-CRF & 14.0 & 14.0 & 41.2 & 15.7 & 4.2 \\
    Two-Step & 15.6 & 15.3 & 43.4 & 17.8 & 8.5 \\
    FEAE & 23.7 & 19.3 & 49.2 & \bf 25.0 & 5.4 \\
    \modelnamebase (Ours) & \bf 24.3 & \bf 21.9 & \bf 49.6 & 24.6 & \bf 11.9 \\
    \midrule
    \midrule
    BART-Gen &  24.3 & 28.1 & 52.4 & 24.8 & 20.8 \\
    \modelnamelarge (Ours) & \bf 28.6 & \bf 30.6 & \bf 53.1 & \bf 27.1 & \bf 22.3 \\
    \bottomrule
    \end{tabular}
}
\caption{
\textbf{Span F1 in RAMS dataset with different sentence distance between trigger and arguments}. 
Most improvements by \modelname come from cross-sentence ($d\neq0)$ arguments extraction.
}
\label{table:cross-rams}
\end{table}
Since there are multiple sentences in the document, some event arguments are located far away from the trigger, which highly increases the difficulty of extraction.
To explore the effect of handling such cross-sentence arguments of our \modelname, we divide the event arguments in RAMS dataset into five bins according to the sentence distance between arguments and trigger, i.e., $d=\left \{ -2, -1, 0, 1, 2 \right \} $.
We report the Span F1 on the RAMS dev set for different methods.
As shown in Table~\ref{table:cross-rams}, the Span F1 for cross-sentence arguments ($d\neq0$) is much lower than local arguments ($d=0$), suggesting the huge challenge to capture long-distance dependency between triggers and cross-sentence arguments.
However, \modelname still surpasses other strong baselines.
In detail, \modelnamebase improves $0.4$ and \modelnamelarge improves $0.7$ F1 compared with the previous state-of-the-art, respectively.
More importantly, when extracting cross-sentence arguments, \modelnamebase and \modelnamelarge yield an improvement of up to $2.3$ and $2.7$ on average.
The results support our claims that \modelname is good at capturing both intra-sentential and inter-sentential features, especially the long-distance between trigger and arguments.

\begin{table}[t]
\centering
\scalebox{0.72}{
    \begin{tabular}{lcccc}
    \toprule
    \multirow{2}{*}{\bf Method} & \multicolumn{2}{c}{\bf Dev} & 
    \multicolumn{2}{c}{\bf Test} \\
    \cmidrule(lr){2-3} \cmidrule(lr){4-5}
    ~ & Span F1 & Head F1 & Span F1 & Head F1 \\
    \midrule
    \modelnamelarge & 49.23 & 56.76 & 51.18 & 58.53 \\
    - Global Encoder & 46.71 & 54.26 & 48.21 & 55.49 \\
    - Local Encoder & 48.43	& 55.44	& 48.69	& 56.82 \\
    - AMR-guided Graph & 48.63 & 55.24 & 49.21 & 56.70 \\
    - Boundary Loss & 47.93	& 55.14	& 50.47	& 57.75 \\
    \bottomrule
    \end{tabular}
}
\caption{
\textbf{Ablation study on RAMS for \modelnamelarge}.
The score would decrease without any kind of module.
}
\label{table:ablaition-rams-large}
\end{table}

\begin{figure*}[htbp]
    \centering
    \includegraphics[width=\linewidth]{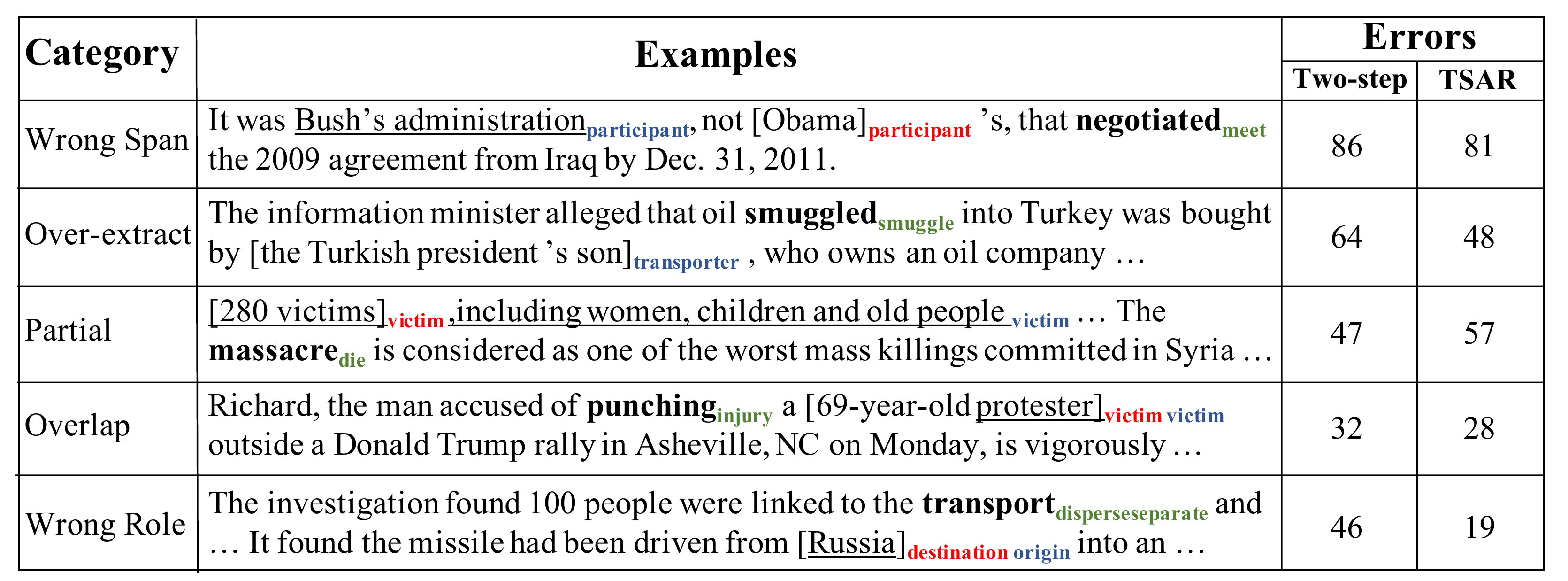}
    \caption{
    \textbf{Error analysis on RAMS dataset}. 
    The triggers are in \textbf{bold} with corresponding event types in green.
    The \underline{underlined} spans refer to golden arguments, with their roles in blue.
    The [bracketed] spans denote the predicted arguments, with their roles noted in red.
    We illustrate the number of different kinds of errors for Two-step and our \modelname, which has $275$ and $233$ errors in total, respectively.
    Compared with Two-step, \modelname decreases errors in most error categories, especially for \emph{Wrong Role} and \emph{Over-extract}.
    }
\label{fig:error-analysis}
\end{figure*}
\begin{figure}[htbp]
    \centering
    \includegraphics[width=\linewidth]{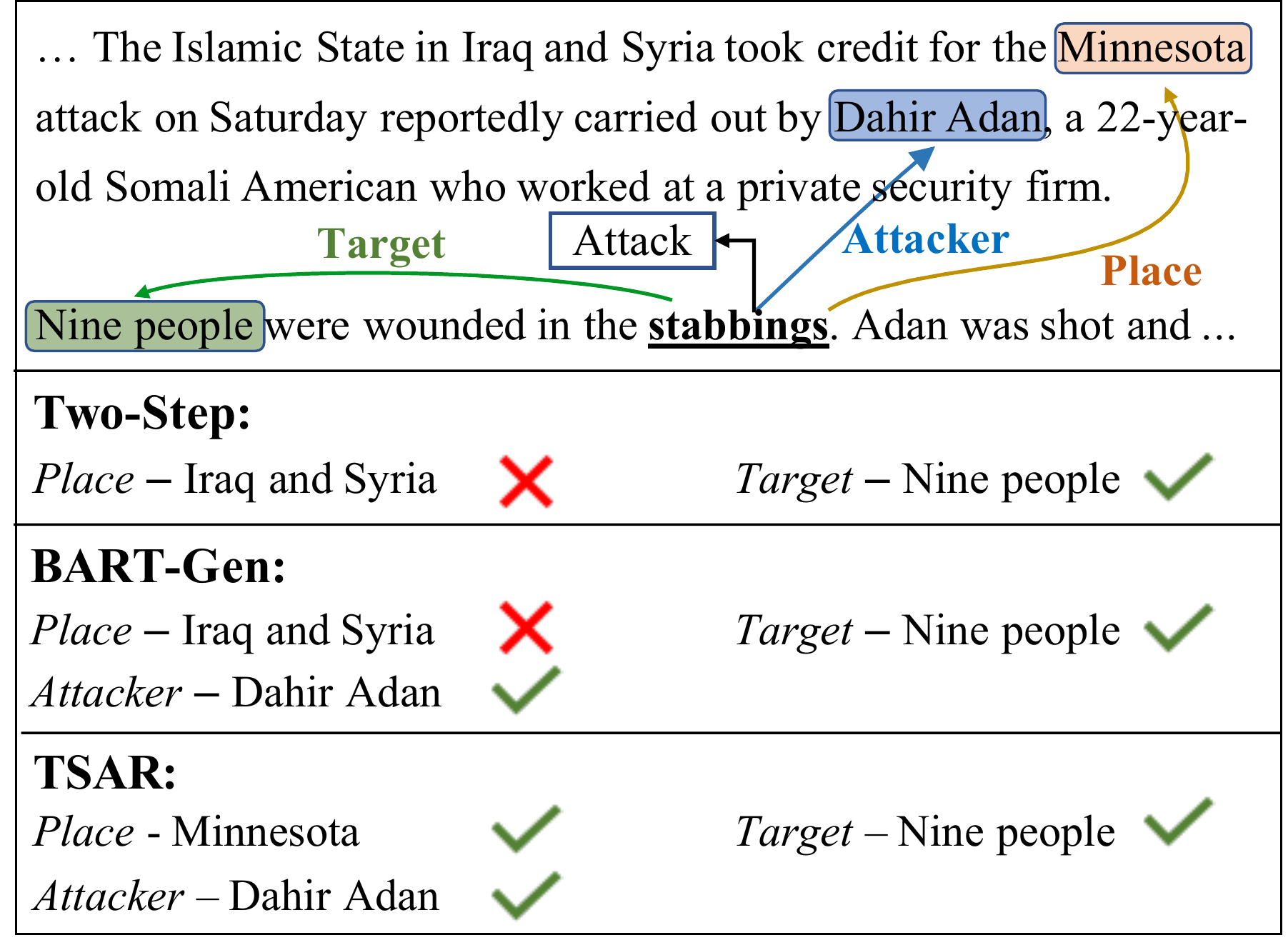}
    \caption{
    \textbf{An extraction case}, where an \emph{Attack} event is triggered by \underline{\textbf{stabbings}} with three arguments. 
    \modelname manages to extract the cross-sentence argument \emph{Minnesota} far from the trigger, while other methods fail.
    }
\label{fig:case-study}
\end{figure}

\subsection{Ablation Study}

We conduct an ablation study to explore the effectiveness of different modules in \modelname.
Table~\ref{table:ablaition-rams-large} show the results on RAMS datasets for \modelnamelarge.
We also provide results for \modelnamebase, and those on WikiEvents datasets in Appendix~\ref{sec:appendix-ablation}.

Firstly, we remove the global or local encoder in the two-stream encoding module.
As shown in Table~\ref{table:ablaition-rams-large}, the removal causes drop in performance, e.g., $3.04$ and $1.71$ Head F1 drop on the test set without global and local encoder.
It suggests the global and local encoders are complementary to each other, and both of them are necessary for \modelname.

Secondly, once we remove the AMR-guided interaction module, the Head F1 would decrease by $1.83$ on the test set.
It shows the semantic structure provided by AMR graphs is helpful to the arguments extraction of the document.

Finally, the removal of boundary loss causes the boundary information lost in span representations, which also leads to $1.62$ and $0.78$ Head F1 decrease on dev and test set.

\subsection{Case Study}

In this section, we show a specific case of the extraction results among different methods.
As shown in Figure~\ref{fig:case-study}, \emph{stabbings} triggers an \emph{Attack} event with three arguments in color.
Since \emph{Nine people} is located near the trigger, all the methods correctly predict it as the \emph{target}.
However, extracting \emph{Minnesota} and \emph{Dahir Adan} asks for capturing long-distance dependency.
Although Two-Step and BART-Gen wrongly predict the \emph{place} as \emph{Iraq and Syria}, and Two-Step even fails to extract the \emph{Attacker}, \modelname manage to extract the cross-sentence arguments.
It can be attributed to that our AMR-enhanced module catches \emph{Minnesota} is the \emph{place} of \emph{attack} that is highly related to the trigger \emph{stabbings} in semantics.

\subsection{Error Analysis}

To further explore the errors made by different models and analyze the reasons in detail, we randomly choose $200$ examples from the RAMS test set and compare the predictions with golden annotations manually.
We divide the errors into five categories, which is shown in Figure~\ref{fig:error-analysis}.
\emph{Wrong Span} refers to assigning a specific role to a wrong span non-overlapped with the golden one.
We find it is usually due to the negative words like \emph{not}, and the coreference spans for the golden one.
\emph{Over-extract} denotes the model predicts an argument role while it does not exist in the document.
Some extracted spans are the sub-strings of the golden spans (\emph{Partial}), or have some overlaps with them (\emph{Overlap}).
These two kinds of errors are usually attributed to the annotation inconsistency in the dataset, such as whether the adjective, quantifier, and article (e.g., \emph{a} and \emph{the}) before the noun should belong to the golden argument.
Besides, the \emph{Partial} error also usually occurs in cases where there is punctuation like a comma in the golden span as shown in Figure~\ref{fig:error-analysis}.
Finally, though the model succeeds to identify the golden span, it can still assign wrong argument role to the span (\emph{Wrong Role}).
We compare the errors of Two-step$_{\mathrm{TCD}}$ and \modelnamebase.
We observe \modelname decrease the number of errors from $275$ to $233$, especially for \emph{Wrong Role} and \emph{Over-extract}, with $27$ and $16$ errors reduction, respectively.

\section{Conclusion}
It is challenging to extract event arguments from a whole document, owing to the long-distance dependency between trigger and arguments over sentences and the distracting context.
To tackle these problems, we propose \textbf{T}wo-\textbf{S}tream \textbf{A}M\textbf{R}-enhanced extraction model (\textbf{\modelname}).
\modelname uses two-stream encoders to encode the document from different perspectives, followed by an AMR-guided interaction module to facilitate the document-level semantic interactions.
An auxiliary boundary loss is introduced to enhance the boundary information for spans.
Experiments on RAMS and WikiEvents datasets demonstrate that \modelname outperform previous state-of-the-art methods by a large margin, with $2.51$ and $5.13$ F1 improvements respectively, especially for cross-sentence argument extraction.

\section*{Acknowledgements}
This paper is supported by the National Key R\&D Program of China under Grand No.2018AAA0102003,
the National Science Foundation of China under
Grant No.61936012 and 61876004.

\bibliography{custom}
\bibliographystyle{acl_natbib}

\clearpage
\appendix

\section{Abstract Meaning Representation (AMR) Graph}
\label{sec:appendix-amr}
There are many AMR parsing approaches \cite{bevil-spring,fernandez-astudillo-etal-2020-transition, wang2021hierarchical, chen-amr}.
To obtain AMR semantic graphs with the align information between text spans and AMR nodes, we use the transition-based AMR parser proposed by~\citet{fernandez-astudillo-etal-2020-transition}, which is a state-of-the-art AMR parser and can achieve satisfactory results for downstream application (up to $81.3$ Smatch on AMR2.0 data).
As the number of AMR relation types is large, which results in too many demanded parameters, we follow~\citet{zhangzixuan} to cluster the relation types into main categories as shown in Table~\ref{table:amr-relations}.
\begin{table}[ht]
\centering
\scalebox{0.8}{
    \begin{tabular}{lc}
    \toprule
    \bf Categories & \bf Relation Types \\
    \midrule
    Spatial & location, destination, path \\
    Temporal & year, time, duration, decade, weekday \\
    Means & instrument, manner, topic, medium \\
    Modifiers & mod, poss \\
    Operators & op-X \\
    Prepositions & prep-X \\
    Sentence & snt \\
    Core Roles & ARG0, ARG1, ARG2, ARG3, ARG4 \\
    Others & other relation types \\
    \bottomrule
    \end{tabular}
}
\caption{
Similar AMR relation types are clustered into the same relation category.
The exception is that ARGx is still treated as an individual relation type.
}
\label{table:amr-relations}
\end{table}

\section{Hyperparameters Setting}
\label{sec:appendix-hyperparameters}
We set the dropout rate to $0.1$, batch size to $8$, and train \modelname  using Adam~\cite{adam} as optimizer with $3\text{e-}5$ learning rate.
We train \modelname for $50$ epochs for RAMS dataset and $100$ epochs for WikiEvents dataset. 
We search the boundary loss weight $\lambda$ from $\left \{  0.05, 0.1, 0.2\right \}$, and $L$ from $\left \{  3, 4 \right \}$, and select the best model using dev set.
Our code is based on Transformers~\cite{transformers} and DGL libraries~\cite{dgl}.

\section{Ablation Study}
\label{sec:appendix-ablation}

In the main body of the paper, we illustrate the results of the ablation study for \modelnamelarge on RAMS dataset.
To thoroughly show the effect of different modules of \modelname, we also provide the results of the ablation study for \modelnamebase on RAMS dataset.
Table~\ref{table:ablaition-rams-base} shows the results on RAMS dataset, from which we can observe removing different modules would cause $1.34\sim2.77$ Span F1 on test set.

Besides, we do ablation study on WikiEvents.
As shown in Table~\ref{table:ablation-wikievent-base}, the Head F1 decreases by $0.70\sim2.02$ and $0.88\sim2.96$ for Arg Identification and Arg Classification sub-tasks respectively, once different modules are removed from \modelnamebase.
Similar conclusions can be drawn from the results of \modelnamelarge, which is shown in Table~\ref{table:ablation-wikievent-large}.

\begin{table}[htbp]
\centering
\scalebox{0.72}{
    \begin{tabular}{lcccc}
    \toprule
    \multirow{2}{*}{\bf Method} & \multicolumn{2}{c}{\bf Dev} & 
    \multicolumn{2}{c}{\bf Test} \\
    \cmidrule(lr){2-3} \cmidrule(lr){4-5}
    ~ & Span F1 & Head F1 & Span F1 & Head F1 \\
    \midrule
    \modelnamebase & 45.23 & 51.70 & 48.06 & 55.04 \\
    - Global Encoder & 43.05 & 50.90 & 45.29 & 53.62 \\
    - Local Encoder & 44.63 & 51.34	& 46.50 & 53.26 \\
    - AMR-guided Graph & 43.57 & 50.80 & 45.97 & 52.85 \\
    - Boundary Loss & 44.42 & 51.08 & 46.72 & 53.91 \\
    \bottomrule
    \end{tabular}
}
\caption{
\textbf{Ablation study on RAMS for \modelnamebase}.
The score would decrease without any kind of module.
}
\label{table:ablaition-rams-base}
\end{table}
\begin{table}[htbp]
\centering
\scalebox{0.72}{
    \begin{tabular}{lcccc}
    \toprule
    \multirow{2}{*}{\bf Method} & \multicolumn{2}{c}{\bf Arg Identification} & 
    \multicolumn{2}{c}{\bf Arg Classification} \\
    \cmidrule(lr){2-3} \cmidrule(lr){4-5}
    ~ & Head F1 & Coref F1 & Head F1 & Coref F1 \\
    \midrule
    \modelnamebase & 75.52 & 73.17 & 68.11 & 66.31 \\
    - Global Encoder & 73.50 & 72.23 & 65.15 & 64.07 \\
    - Local Encoder & 74.40 & 72.62 & 67.11 & 65.41 \\
    - AMR-guided Graph & 73.88 & 72.45 & 65.83 & 64.94 \\
    - Boundary Loss & 74.82 & 72.50 & 67.23 & 65.95 \\
    \bottomrule
    \end{tabular}
}
\caption{
\textbf{Ablation study on WikiEvents for \modelnamebase}.
The performance of identification and classification would decrease without any kind of module.
}
\label{table:ablation-wikievent-base}
\end{table}
\begin{table}[htbp]
\centering
\scalebox{0.72}{
    \begin{tabular}{lcccc}
    \toprule
    \multirow{2}{*}{\bf Method} & \multicolumn{2}{c}{\bf Arg Identification} & 
    \multicolumn{2}{c}{\bf Arg Classification} \\
    \cmidrule(lr){2-3} \cmidrule(lr){4-5}
    ~ & Head F1 & Coref F1 & Head F1 & Coref F1 \\
    \midrule
    \modelnamelarge & 76.62 & 75.52 & 69.70 & 68.79 \\
    - Global Encoder & 74.12 & 72.80 & 67.54 & 66.41 \\
    - Local Encoder & 74.60 & 73.32 & 68.08 & 66.88 \\
    - AMR-guided Graph & 74.52 & 73.82 & 67.67 & 66.54 \\
    - Boundary Loss & 75.50 & 74.05 & 68.60 & 67.33 \\
    \bottomrule
    \end{tabular}
}
\caption{
\textbf{Ablation study on WikiEvents for \modelnamelarge}.
The performance of identification and classification would decrease without any kind of module.
}
\label{table:ablation-wikievent-large}
\end{table}

\end{document}